%% file: GlobalSIP16_MSGC_V3.tex
\documentclass{article}
\usepackage{spconf}
\usepackage{url}
\usepackage{graphicx}
\usepackage{epstopdf}
\usepackage{algorithm}
\usepackage{algorithmicx}
\usepackage{algpseudocode}
\usepackage{cite}
\usepackage{color}
\usepackage{amsmath,amssymb,amsfonts,amsthm}
\usepackage{bm}
\usepackage{setspace}
\usepackage{multirow}
\usepackage{mathtools}
\usepackage{wrapfig}
\usepackage{subcaption}

\newtheorem{thm} {Theorem}

\include{notation_20160812}

\begin{document}
\title{Multilayer Spectral Graph Clustering via Convex Layer Aggregation}

\name{Pin-Yu Chen
	\thanks{This work was partially supported by the US Army Research Office, grant W911NF-15-1-0479.}
	 \qquad Alfred O. Hero III,~Fellow,~IEEE	
	}
\address{Department of Electrical Engineering and Computer Science, University of Michigan, Ann Arbor, USA \\ \{pinyu, hero\}@umich.edu 
}




\ninept 
\maketitle
\begin{abstract}
Multilayer graphs are commonly used for representing different relations between entities and handling heterogeneous data processing tasks. New challenges arise in multilayer graph clustering for assigning clusters to a common multilayer node set and for combining information from each layer.  This paper presents a theoretical framework for multilayer spectral graph clustering of the nodes via convex layer aggregation. Under a novel multilayer signal plus noise model, we provide a phase transition analysis that establishes the existence of a critical value on the noise level that permits reliable cluster separation. The analysis also specifies analytical upper and lower bounds on the critical value, where the bounds become exact when the clusters have identical sizes. Numerical experiments on synthetic multilayer graphs are conducted to validate the phase transition analysis and study the effect of layer weights and noise levels  on clustering reliability.
\end{abstract}

\section{Introduction}

Multilayer graphs are useful for representing different relations between entities and handing heterogeneous multilayer data processing tasks, where each layer describes a specific type of connectivity pattern among a common node set across layers. For example, in multi-relational social networks, each layer corresponds to one type of social relation. In temporal networks, each layer corresponds to the snapshot of the entire network at a sampled time instance. Multilayer graphs have been applied to many signal processing and data mining techniques, including inference of mixture models \cite{Oselio14,xu2014dynamic}, tensor decomposition \cite{Domenico13tensor}, information extraction \cite{oselio2015information}, multi-view learning and processing \cite{xiaowen2014multi}, graph wavelet transform \cite{leonardi2013tight}, principal component analysis and dictionary learning \cite{benzi2016principal,CPY16ICASSP}, anomaly detection \cite{park2013anomaly}, and community detection \cite{kivela2014multilayer,kim2015community,CPY14deep}, among others.

In particular, the task of multilayer graph clustering is to find a consensus cluster assignment on each node in the common node set by inspecting the connectivity pattern in each layer. Different from clustering in single-layer graphs, clustering in multilayer graph faces new challenges due to (1) information aggregation from multiple layers, and (2) lack of a theoretical framework on clustering reliability assessment. By viewing the connectivity pattern in each layer as a signal plus noise model, this paper aims to provide a theoretical framework for analyzing the performance of multilayer spectral graph clustering (SGC) via convex layer aggregation, where spectral clustering is implemented on an aggregated graph via convex combination of each layer.
Specifically, fixing the within-cluster edges (signals) and varying the parameters governing the between-cluster edges (noises), we show that the accuracy of multilayer SGC can be separated into two regimes: a reliable regime where high clustering accuracy can be guaranteed, and an unreliable regime where  high clustering accuracy is impossible. Moreover, we show that the upper and lower bounds on the critical noise level that separates these two regimes are closed-form functions of  the signal strength, the number of clusters, the cluster size distributions, and the layer weight vector for convex layer aggregation. In addition, the bounds become exact in the case of identical cluster sizes. Numerical experiments on synthetic multilayer graphs are conducted to validate the phase transition analysis and study the effect of layer weights and noise levels  on clustering reliability.

\section{Related Work}
\label{sec_related}
Layer aggregation has been a principal method for processing and mining multilayer graphs \cite{cai2005community,tang2009uncoverning,wu2015discovering,tang2012community,de2015structural,Taylor16}, as it transforms a multilayer graph into a single aggregated graph,  facilitating application of data analysis techniques designed for single-layer graphs.
Extending from the stochastic block model (SBM) for graph clustering in single-layer graphs \cite{Holland83,CPY14modularity}, multilayer SBM has been proposed for graph clustering on multilayer graphs \cite{han2015consistent,paul2015community,barbillon2016stochastic,PhysRevX.6.011036,Taylor16,Stanley16}. Under the assumption of two equally-sized clusters, the authors in \cite{Taylor16} show that if each layer is an independent realization of a common SBM,  the inferential limit for cluster detectability decays with $O(L^{-\frac{1}{2}})$, where $L$ is the number of layers. In \cite{Stanley16}, a layer selection method based on a multilayer SBM is proposed to improve the performance of graph clustering. However, the multilayer SBM assumes homogeneous connectivity structure for within-cluster and between-cluster edges in each layer, and it assumes layer-wise independence.
 The multilayer signal plus noise model considered in this paper is a general model that includes the multilayer SBM, as it does not impose any distributional assumption on the within-cluster connectivity for each layer. More details on multilayer graph models for graph clustering can be found in the recent survey papers \cite{kivela2014multilayer,kim2015community}.

\section{Multilayer Signal Plus Noise Model}
We consider the multilayer graph model of $L$ layers representing different relationships among a common node set $\cV$ of $n$ nodes. The graph in the $\ell$-th layer is an undirected graph with nonnegative edge wights, which is denoted by $G_\ell=(\cV,\cE_{\ell})$, where $\cE_{\ell}$ is the set of weighted edges in the $\ell$-th layer. The $n \times n$ binary symmetric adjacency matrix $\bAl$ is used to represent the connectivity structure of $G_\ell$. The entry $[\bAl]_{uv}=1$ if nodes $u$ and $v$ are connected in the $\ell$-th layer, and $[\bAl]_{uv}=0$ otherwise. Similarly, the $n \times n$ nonnegative symmetric weight matrix $\bWl$ is used to represent the edge weights in $G_\ell$, where $\bWl$ and $\bAl$ have the same zero structure.

We assume each layer in the multilayer graph is a (possibly correlated) representation of common $K$ clusters that partitions the node set $\cV$, where the $k$-th cluster has cluster size $n_k$ such that $\sum_{k=1}^K n_k=n$. 
$\nmin=\min_{k \in \{1,\ldots,K\}} n_k$ and $\nmax=\max_{k \in \{1,\ldots,K\}} n_k$ denote the largest and smallest cluster size, respectively.
Specifically, the adjacency matrix $\bAl$ of $G_\ell$ in the $\ell$-th layer can be represented as 
\begin{align}                                                              \label{eqn_network_model_multilayer}
\bAl= \begin{bmatrix}
\bAl_1          & \bCl_{12} & \bCl_{13} & \cdots & \bCl_{1K}           \\
\bCl_{21}       & \bAl_2    & \bCl_{23} & \cdots & \bCl_{2K} \\
\vdots         & \vdots   & \ddots   & \vdots  & \vdots  \\
\vdots         & \vdots   & \vdots   & \ddots  & \vdots  \\
\bCl_{K1}       & \bCl_{K2} & \cdots   & \cdots  & \bAl_{K}
\end{bmatrix},
\end{align}
where $\bAl_k$ is an $n_k \times n_k$ binary symmetric matrix denoting the adjacency matrix of within-cluster edges of the $k$-th cluster in the $\ell$-th layer, and $\bCl_{ij}$  is an $n_i \times n_j$ binary rectangular matrix denoting the adjacency matrix of between-cluster edges of clusters $i$ and $j$  in the $\ell$-th layer, $1 \leq i,j \leq K$, $i \neq j$, 
and $\bCl_{i j}={\bCl_{ij}}^T$.

Similarly, the edge weight matrix $\bWl$  can be represented as 
\begin{align}                                                              \label{eqn_network_model_multilayer_weight}
\bWl= \begin{bmatrix}
\bWl_1          & \bFl_{12} & \bFl_{13} & \cdots & \bFl_{1K}           \\
\bFl_{21}       & \bWl_2    & \bFl_{23} & \cdots & \bFl_{2K} \\
\vdots         & \vdots   & \ddots   & \vdots  & \vdots  \\
\vdots         & \vdots   & \vdots   & \ddots  & \vdots  \\
\bFl_{K1}       & \bFl_{K2} & \cdots   & \cdots  & \bWl_{K}
\end{bmatrix},
\end{align}
where  $\bWl_k$ is an $n_k \times n_k$ nonnegative symmetric matrix denoting the edge weights of within-cluster edges of the $k$-th cluster in the $\ell$-th layer, and $\bFl_{ij}$  is an $n_i \times n_j$ nonnegative rectangular matrix denoting the edge weights  of between-cluster edges of clusters $i$ and $j$  in the $\ell$-th layer, $1 \leq i,j \leq K$, $i \neq j$, and $\bFl_{ij}={\bFl_{ij}}^T$.

Using the cluster-wise block representations of the adjacency and edge weight matrices for the multilayer graph model described in (\ref{eqn_network_model_multilayer}) and  (\ref{eqn_network_model_multilayer_weight}), we propose a signal plus noise model for $\bAl$ and $\bWl$ to analyze the effect of convex layer aggregation on graph clustering. 
Specifically, for each layer we assume the connectivity structure and edge weight distributions follow the random interconnection model (RIM)  \cite{CPY16AMOS}. 
In RIM the signal of the $k$-th cluster in the $\ell$-th layer is the connectivity structure and weights of the within-cluster edges represented by the matrices $\bAl_k$ and $\bWl_k$, respectively. In particular, analogous to the formulation of many detection problems in signal processing, the signal can be arbitrary in the sense that we impose no distributional assumption for the within-cluster edges. The noise between clusters $i$ and $j$ in the $\ell$-th layer is the connectivity structure and weights of the between-cluster edges represented by the matrices $\bCl_{ij}$ and $\bFl_{ij}$, respectively.

Throughout this paper, we assume the connectivity of a between-cluster edge (i.e., the noise) in each layer is independently drawn from a 
layer-wise and block-wise independent common Bernoulli distribution. Specifically, each entry in $\bCl_{ij}$ representing the existence of an edge between clusters $i$ and $j$ in the $\ell$-layer 
is an independent realization of a Bernoulli random variable with edge connection probability $\pijl \in [0,1]$ that is layer-wise and block-wise independent. In addition, given the existence of an edge $(u,v)$ between clusters $i$ and $j$ in the $\ell$-layer,
the entry $[\bFl_{ij}]_{uv}$ representing the corresponding edge weight is independently drawn from a nonnegative distribution with mean $\Wbarijl$ and bounded fourth moment that is  layer-wise and block-wise independent.

For the $\ell$-th layer, the noise accounting for the between-cluster edges is said to be \textit{block-wise identical} if the noise parameters $\pijl=\pl$ and $\Wbarijl=\Wbarl$ for every cluster pair $i$ and $j$, $i \neq j$. Otherwise it is said to be \textit{block-wise non-identical}.

\section{Multilayer Spectral Graph Clustering via Convex Layer Aggregation}

\subsection{Notations and mathematical formulations}
Let $\bw=[w_1,\ldots,w_L]^T \in \cW_{L}$ be an $L \times 1$ column vector representing the layer weight vector for convex layer aggregation, where $\cW_{L}=\{\bw: w_\ell \geq 0,~\sum_{\ell=1}^{L} w_\ell=1\}$ is the set of feasible layer weight vectors. The single-layer graph obtained via convex layer aggregation with layer weight vector $\bw$ is denoted by $G^{\bw}$. The (weighted) adjacency matrix $\bAw$ and the edge weight matrix $\bWw$ of $G^{\bw}$ satisfy $\bAw=\sum_{\ell=1}^{L} w_\ell \bAl$ and $\bWw=\sum_{\ell=1}^{L} w_\ell \bWl$. The graph Laplacian matrix $\bLw$ of $G^\bw$ is defined as  $\bLw=\bSw-\bWw=\sum_{\ell=1}^{L} w_\ell \bLl$, where $\bSw=\diag(\bsw)$ is a diagonal matrix, $\bsw=\bWw \bone_n$ is the vector of nodal strength of $G^\bw$,  $\bone_n$ is the $n \times 1$ column vector of ones, and $\bLl$ is the graph Laplacian matrix of $G_\ell$. Similarly, the graph Laplacian matrix $\bL_k^\bw$ accounting for the within-cluster edges of the $k$-th cluster in $G^\bw$ is defined as 
$\bL_k^\bw=\bSw_k-\bWw_k=\sum_{\ell=1}^{L} w_\ell \bLl_k$, where  $\bWw_k=\sum_{\ell=1}^{L} w_\ell \bWl_k$, $\bSw_k=\diag(\bWw_k\bone_{n_k})$, and $\bLl_k=\bSl_k-\bWl_k$. The $i$-th smallest eigenvalue of $\bLw$ is denoted by $\lambda_i(\bLw)$. Based on the definition of $\bLw$, the smallest eigenvalue $\lambda_1(\bLw)$ of $\bLw$ is 0, since $\bLw \bone_n=\bzero_n$, where $\bzero_n$ is the $n \times 1$ column vector of zeros.

Spectral graph clustering (SGC) \cite{Luxburg07} partitions the nodes in $G^\bw$ 
into $K$ ($K \geq 2$) clusters based on the $K$ eigenvectors associated with the $K$ smallest eigenvalues of $\bLw$. Specifically, SGC first transforms a node in $G^\bw$ to a $K$-dimensional vector in the subspace spanned by these eigenvectors, and then implements K-means clustering \cite{hartigan1979algorithm} on the $K$-dimensional vector space representation to group the nodes in $G^\bw$ into $K$ clusters based on their distances. For analysis purposes, throughout this paper we assume $G^\bw$ is a connected graph. If  $G^\bw$ is connected, it is known that $\lambda_i(\bLw)>0$ for all $i \geq 2$ \cite{Fiedler73}. Furthermore, the eigenvector associated with the smallest eigenvalue $\lambda_1(\bLw)$ provides no information about graph clustering since it is proportional to $\bone_n$.

Let $\bY \in \mathbb{R}^{n \times (K-1)}$ denote the eigenvector matrix where its $k$-th column is the $(k+1)$-th eigenvector associated with $\lambda_{k+1}(\bLw)$, $1 \leq k \leq K-1$. By the Courant-Fischer theorem \cite{jennings1992matrix}, $\bY$ is the solution to the minimization problem 
\begin{align}
\label{eqn_spectral_clustering_multi_formulation_ML}
&\SK(\bLw)=\min_{\bX \in \mathbb{R}^{n \times (K-1)}} \trace(\bX^T \bLw \bX), \nonumber \\
&\text{subjec~to}~\bX^T \bX= \bI_{K-1},~\bX^T \bone_n=\bzero_{K-1}, 
\end{align}
where the optimal value $\SK(\bLw)=\trace(\bY^T \bLw \bY)$ in (\ref{eqn_spectral_clustering_multi_formulation_ML}) is the partial eigenvalue sum $\SK(\bLw)=\sum_{k=2}^{K} \lambda_k(\bLw)$, $\bI_{K-1}$ is the $(K-1) \times (K-1)$ identity matrix, and the constraints in  (\ref{eqn_spectral_clustering_multi_formulation_ML}) impose  orthonormality and centrality on the eigenvectors. In summary, multilayer SGC via convex layer aggregation works by computing the eigenvector matrix $\bY$ from $\bLw$ of $G^\bw$, and implementing K-means clustering on the rows of $\bY$ to group the nodes into $K$ clusters.

\subsection{Phase transitions  under block-wise identical noise}

Under the multilayer signal plus noise model, if we further assume block-wise identical noise, then the noise level in the $\ell$-th layer can be characterized by the parameter $\tl=\pl \cdot \Wbarl$, where $\pl \in [0,1]$ is the edge connection parameter and $\Wbarl>0$ is the mean of the between-cluster edge weights in the $\ell$-th layer.
Given a layer weight vector $\bw \in \cW_{L}$, let $\tw=\sum_{\ell=1}^{L} w_\ell \tl$ denote the aggregated noise level of the graph $G^\bw$.
Theorem \ref{thm_spec_ML} below establishes phase transitions in the eigendecomposition of the graph Laplacian matrix $\bLw$ of $G^\bw$. We show that there exists a critical value $\twstar$ such that the $K$ smallest eigenpairs of  $\bLw$ that are used for multilayer SGC have different characteristics when $\tw < \twstar$ and $\tw > \twstar$.
In particular, we show that the solution to the minimization problem in (\ref{eqn_spectral_clustering_multi_formulation_ML}), the eigenvector matrix $\bY=[\bY_1^T,\bY_2^T,\ldots,\bY_K^T]^T \in \bbR^{n \times (K-1)}$ represented by the cluster partitioned form, where $\bY_k \in \bbR^{n_k \times (K-1)}$ with its rows indexing the nodes in cluster $k$, has cluster-wise separability when  $\tw < \twstar$
in the sense that the matrices $\{\bY_k\}_{k=1}^K$ are row-wise identical and cluster-wise distinct, whereas when  $\tw >  \twstar$ the row-wise average of each matrix $\bY_k$ is a zero vector and hence the clusters are not separable by inspecting the rows of  $\bY$.

\begin{thm}[block-wise identical noise]~\\
	\label{thm_spec_ML}
 Given a layer weight vector $\bw \in \cW_L$, and assuming the block-wise identical noise model with aggregated noise level $\tw=\sum_{\ell=1}^{L} w_\ell \tl$, let $\cwstar=\min_{k \in \{1,2,\ldots,K\}} \LB \frac{\SK(\bLw_k)}{n}  \RB$, where $\bLw_k=\sum_{\ell=1}^{L} w_\ell \bLl_k$.
There exists a critical value $\twstar$ such that the following holds almost surely as $n_k \ra \infty$~$\forall~k$ and $\frac{\nmin}{\nmax} \ra c >0$: \\	
	\textnormal{(a)}~$ \left\{
	\begin{array}{ll}
	\textnormal{If~} \tw \leq \twstar,~ \frac{\SK(\bLw)}{n} = (K-1)\tw; \\
	\textnormal{If~} \tw > \twstar,~ \cwstar + (K-1) \lb 1-\frac{\nmax}{n} \rb \tw  \leq  	\frac{\SK(\bLw)}{n} \\ 
	~~~~~~~~~~~~~~~~~~~~~~~\leq \cwstar + (K-1) \lb 1-\frac{\nmin}{n} \rb \tw ;  \\
	\textnormal{If~} \tw > \twstar \textnormal{~and~} c=1,~ \frac{\SK(\bLw)}{n} = \cwstar +\frac{(K-1)^2}{K} \tw.
	\end{array}
	\right.$ \\
	Furthermore, \\	
	\textnormal{(b)}~$\left\{	
	\begin{array}{ll}
	\textnormal{If~} \tw < \twstar,~\bY_k = \bone_{n_k} \bone_{K-1}^T \bV_k\\
	~~~~~~~~~~~~~~~~~~~~~~~~~~~~=\Lb v^k_1 \bone_{n_k},v^k_2 \bone_{n_k},\ldots,v^k_{K-1} \bone_{n_k} \Rb,~\forall~k; \\
	\textnormal{If~} \tw > \twstar,~
	\bY_k^T\bone_{n_k} = \bzero_{K-1},~\forall~k; \\
	\textnormal{If~} \tw = \twstar,~\bY_k =\bone_{n_k} \bone_{K-1}^T \bV_k \textnormal{~or~} \bY_k^T\bone_{n_k} = \bzero_{K-1}, ~\forall~k,
	\end{array}
	\right.$ \\
	where $\bV_k=\diag(v^k_1, v^k_2,\ldots, v^k_{K-1}) \in \mathbb{R}^{(K-1) \times (K-1)}$. \\
	In particular, when $\tw < \twstar$, $\bY$ has the following properties:\\
	\textnormal{(b-1)} The columns of $\bY_k$ are constant vectors. \\
	\textnormal{(b-2)} Each column of $\bY$ has at least two nonzero cluster-wise constant components, and these constants have alternating signs such that their weighted sum equals $0$ (i.e., $\sum_{k} n_k v^k_j = 0,~\forall~j \in\{1,2,\ldots,K-1\}$). \\
	\textnormal{(b-3)} No two columns of $\bY$ have the same sign on the cluster-wise nonzero components.	 \\	
	Finally, $\twstar$ satisfies: \\	
	\textnormal{(c)}~$\tLBw \leq \twstar \leq \tUBw$, where \\
$	\tLBw = \frac{\min_{k \in \{1,2,\ldots,K\}} \SK(\bLw_k)}{(K-1)\nmax}; $~
$	\tUBw  = \frac{\min_{k \in \{1,2,\ldots,K\}} \SK(\bLw_k)}{(K-1)\nmin}.$
 \\
	In particular, 	$\tLBw=\tUBw$ when $c=1$.	
\end{thm}	

Theorem \ref{thm_spec_ML} (a) establishes a phase transition in the increase
of the normalized partial eigenvalue sum $\frac{\SK(\bLw)}{n}$ with respect to the aggregated noise level $\tw$. When $\tw \leq \twstar$ the quantity $\frac{\SK(\bL)}{n}$ is exactly  $(K-1)\tw$. When $\tw> \twstar$ the slope in $\tw$ of $\frac{\SK(\bL)}{n}$  changes and the intercept $c^*=\min_{k \in \{1,2,\ldots,K\}} \LB \frac{\SK(\bLw_k)}{n}  \RB=\min_{k \in \{1,2,\ldots,K\}} \LB \frac{\sum_{\ell=1}^{L} w_\ell \SK(\bLl_k)}{n}  \RB$ depends on the cluster having the smallest aggregated partial eigenvalue sum given a layer weight vector $\bw$. In particular, when all clusters have the same size (i.e., $\nmax=\nmin=\frac{n}{K}$) so that $c=1$, $\frac{\SK(\bL)}{n}$ undergoes a slope change from $K-1$ to $\frac{(K-1)^2}{K}$ at the critical value $\tw=\twstar$.

Theorem \ref{thm_spec_ML} (b) establishes a phase transition in cluster-wise separability of the eigenvector matrix $\bY$ for multilayer SGC. When $\tw < \twstar$, the conditions (b-1) to (b-3) imply that the rows of the cluster-wise components $\{\bY_k\}_{k=1}^K$ are coherent, and hence the row vectors in $\bY$ possess cluster-wise separability. On the other hand, when $\tw > \twstar$,
the row sum of each $\bY_k$ is a zero vector, making $\mathbf Y_k$ incoherent. This means that the entries of each column in $\bY_k$ have alternating signs and hence K-means clustering on the rows of $\bY$ yields incorrect clusters.

Theorem  \ref{thm_spec_ML} (c) establishes upper and lower bounds on the critical threshold value $\twstar$ of the aggregated noise level $\tw$ given a layer weight vector $\bw$. These bounds are determined by the cluster having the smallest aggregated partial eigenvalue sum $\SK(\bLw_k)=\sum_{\ell=1}^{L} w_\ell \SK(\bLl_k)$, the number of clusters $K$, and the largest and smallest cluster size ($\nmax$ and $\nmin$). When all cluster sizes are identical (i.e., $c=1$), these bounds become tight (i.e., $\tLBw=\tUBw$). Moreover, by the nonnegativity of the layer weights we can obtain a universal lower bound on $\tLBw$ for any $\bw \in \cW_L$, which is 
\begin{align}
\label{eqn_LB_tLB}
\tLBw 
\geq \frac{\min_{k \in \{1,2,\ldots,K\}}  \min_{\ell \in \{1,2,\ldots,L\}}\SK(\bLl_k)}{(K-1)\nmax}. 
\end{align}
Since $\SK(\bLl_k)$ is a measure of connectivity for cluster $k$ in the $\ell$-th layer, the lower bound of $\tLBw$ in (\ref{eqn_LB_tLB}) implies that the performance of multilayer SGC is indeed affected by the least connected cluster among all $K$ clusters and across $L$ layers. Specifically, if the graph in each layer is unweighted and $K=2$, then $\SK(\bLl_k)=\lambda_2(\bLl_k)$ reduces to the algebraic connectivity \cite{Fiedler73,CPY14spectral} of cluster $k$ in the $\ell$-th layer.  

\subsection{Phase transitions under block-wise non-identical noise}
Under the block-wise non-identical noise model, the noise level of between-cluster edges between clusters $i$ and $j$ in the $\ell$-th layer is characterized by the parameter $\tijl=\pijl \cdot \Wbarijl$, $1 \leq i,j \leq K$, $i \neq j$, and $1 \leq \ell \leq L$. Let $\tlmax=\max_{1\leq i,j \leq K,~i \neq j} \tijl$ be the maximum noise level in the $\ell$-th layer and let $\twmax=\sum_{\ell=1}^{L} w_\ell \tlmax$ denote the aggregated maximum noise level given a layer weight vector $\bw \in \cW_L$.

Let $\bY \in \mathbb{R}^{n \times (K-1)}$ be the  eigenvector matrix of $\bLw$ under the
block-wise non-identical noise model, and let  $\btY \in \mathbb{R}^{n \times (K-1)}$ be the eigenvector matrix of the graph Laplacian $\btLw$ of another random graph generated under the block-wise identical noise model with aggregated noise level $\tw$, which is independent of $\bL$.   Theorem \ref{thm_principal_angle_ML} below specifies the distance between the subspaces spanned by the columns of $\bY$ and $\btY$ by inspecting their principal angles \cite{Luxburg07}. Specifically, since $\bY$ and $\btY$ both have orthonormal columns, the vector $\bv$ of $K-1$ principal angles between their column spaces is $\bv=[\cos^{-1}\sigma_1(\bY^T \btY),\ldots,\cos^{-1}\sigma_{K-1}(\bY^T \btY)]^T$, where $\sigma_k(\bM)$ is the $k$-th largest singular value of a real rectangular matrix $\bM$.
Let $\mathbf{\Theta}(\bY,\btY)=\diag(\bv)$, and let $\sin\mathbf{\Theta}(\bY,\btY)$ be defined entrywise. 
When $\tw < \twstar$, Theorem \ref{thm_principal_angle_ML} provides an upper bound on the Frobenius norm of $\sin\mathbf{\Theta}(\bY,\btY)$, which is denoted by $\|\sin\mathbf{\Theta}(\bY,\btY) \|_F$. Moreover, if $\twmax< \twstar$, where $\twstar$ is the critical threshold value for the block-wise identical noise model as specified in Theorem \ref{thm_spec_ML}, then $\| \sin\mathbf{\Theta}(\bY,\btY) \|_F$ can be further bounded.

\begin{thm}[block-wise non-identical noise]~\\
	\label{thm_principal_angle_ML}
	Given a layer weight vector $\bw \in \cW_L$, and assuming the block-wise non-identical noise model with maximum noise level $\{\tlmax\}_{\ell=1}^L$ for each layer, let $\twstar$  be 
	be the critical threshold value for the block-wise identical noise model specified by Theorem \ref{thm_spec_ML}, and define $\delta_{\tw,n}=\min\{\tw,|\lambda_{K+1}(\frac{\bLw}{n})-\tw|\}$. 
	For a fixed $\tw$, if $\tw < \twstar$ and $\delta_{\tw,n} \ra \delta_{\tw} > 0$ as $n_k \ra \infty$~$\forall~k$,
	the following statement holds almost surely as
	$n_k \ra \infty$~$\forall~k$ and $\frac{\nmin}{\nmax} \ra c >0$:
	\begin{align}
	\label{eqn_principal_angle_bound_ML}
	\|\sin\mathbf{\Theta}(\bY,\btY)\|_F \leq \frac{\| \bLw - \btLw \|_F}{n \delta_{\tw}}.
	\end{align}
	Furthermore, let $\twmax=\sum_{\ell=1}^L w_\ell \tlmax$. If $\twmax < \twstar$,
	\begin{align}
	\label{eqn_principal_angle_bound_2_ML}
	\|\sin\mathbf{\Theta}(\bY,\btY)\|_F \leq \min_{\tw \leq \twmax} \frac{\| \bLw - \btLw \|_F}{n \delta_{\tw}}.
	\end{align}
\end{thm}

Theorem \ref{thm_principal_angle_ML} shows that the subspace distance  $\|\sin\mathbf{\Theta}(\bY,\btY)\|_F$ is upper bounded by (\ref{eqn_principal_angle_bound_ML}), where $\btY$ is the eigenvector matrix of $\btLw$ under the block-wise identical noise model when its aggregated noise level $\tw < \twstar$. Furthermore, if the aggregated maximum noise level $\twmax < \twstar$, then a tight upper bound on $\|\sin\mathbf{\Theta}(\bY,\btY)\|_F$ can be obtained by (\ref{eqn_principal_angle_bound_2_ML}).
Therefore, using the cluster-wise separability of $\btY$ as established in Theorem \ref{thm_spec_ML} (b), when $\twmax < \twstar$,
cluster-wise separability in $\bY$ can be expected provided that $\|\sin\mathbf{\Theta}(\bY,\btY)\|_F $ is small. 
The proofs of Theorems $\ref{thm_spec_ML}$ and $\ref{thm_principal_angle_ML}$ are given in the supplementary file.

\section{Numerical Results}

To validate the phase transitions in the accuracy of multilayer SGC via convex layer aggregation, we generate synthetic multilayer graphs from a two-layer correlated multilayer graph model. Specifically, we generate edge connections within and between $K=3$ equally-sized ground-truth clusters on $L=2$ layers $G_1$ and $G_2$. The two layers $G_1$ and $G_2$ are correlated since their edge connections are generated in the following manner.
For every node pair ($u,v$) of the same cluster, with probability $q_{11}$ there is a within-cluster edge ($u,v$) in $G_1$ and $G_2$, with probability  $q_{10}$ there is a within-cluster edge  ($u,v$) in $G_1$ but not in $G_2$, with probability  $q_{01}$ there is a within-cluster edge  ($u,v$) in $G_2$ but not in $G_1$, and with probability  $q_{00}$ there is no edge  ($u,v$) in $G_1$ and $G_2$.
These four parameters are nonnegative and sum to $1$. For between-cluster edges, we adopt the block-wise identical noise model such that for each layer $\ell$, the edge connection between every node pair from different clusters is an i.i.d. Bernoulli random variable with parameter $\pl$.

\subsection{Phase transitions incurred by noise levels}
\begin{figure}[t]
		\centering
		\begin{subfigure}[b]{0.245\linewidth}
			\includegraphics[width=\textwidth]{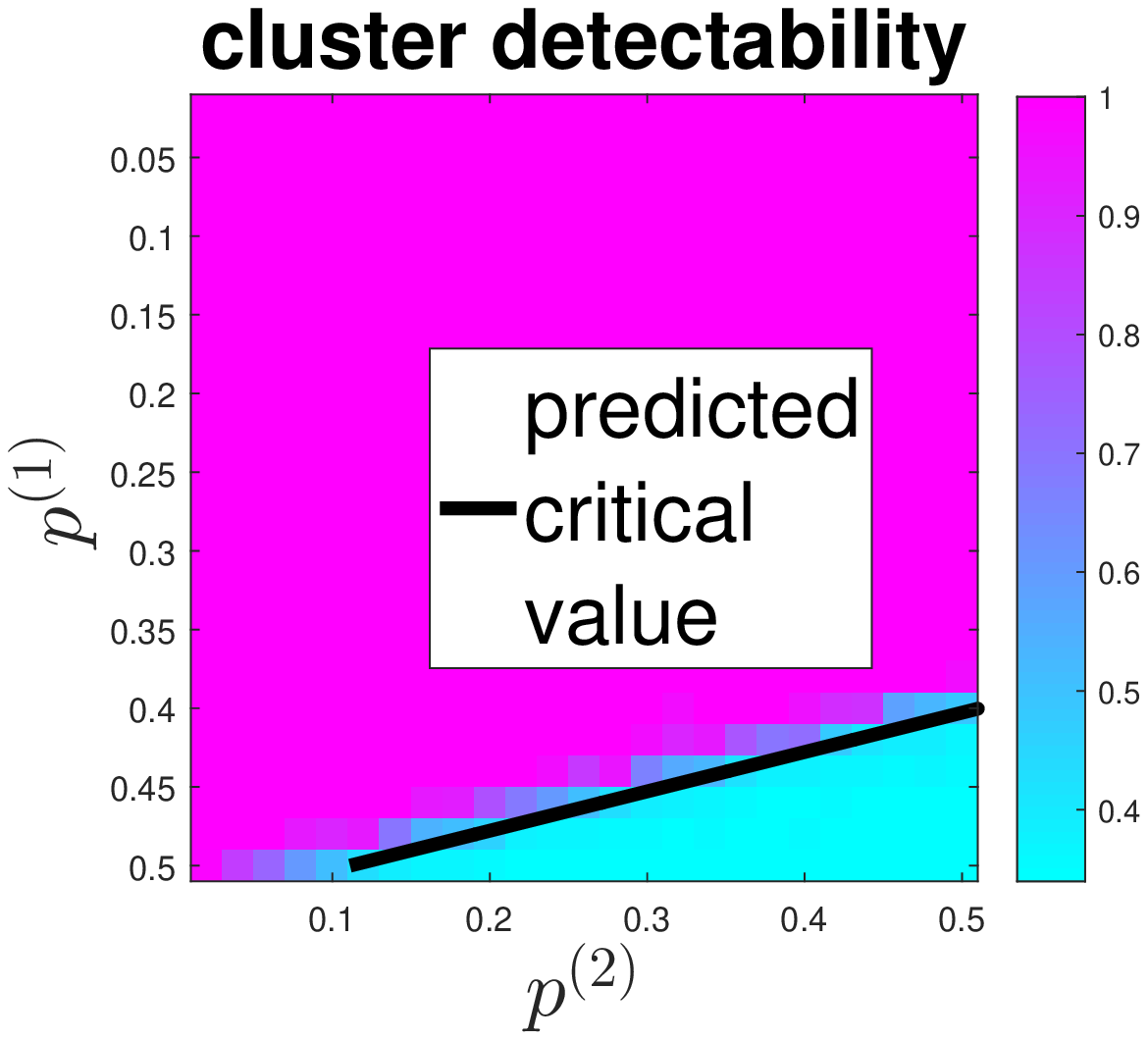}
		\end{subfigure}%
		\centering
		\begin{subfigure}[b]{0.245\linewidth}
			\includegraphics[width=\textwidth]{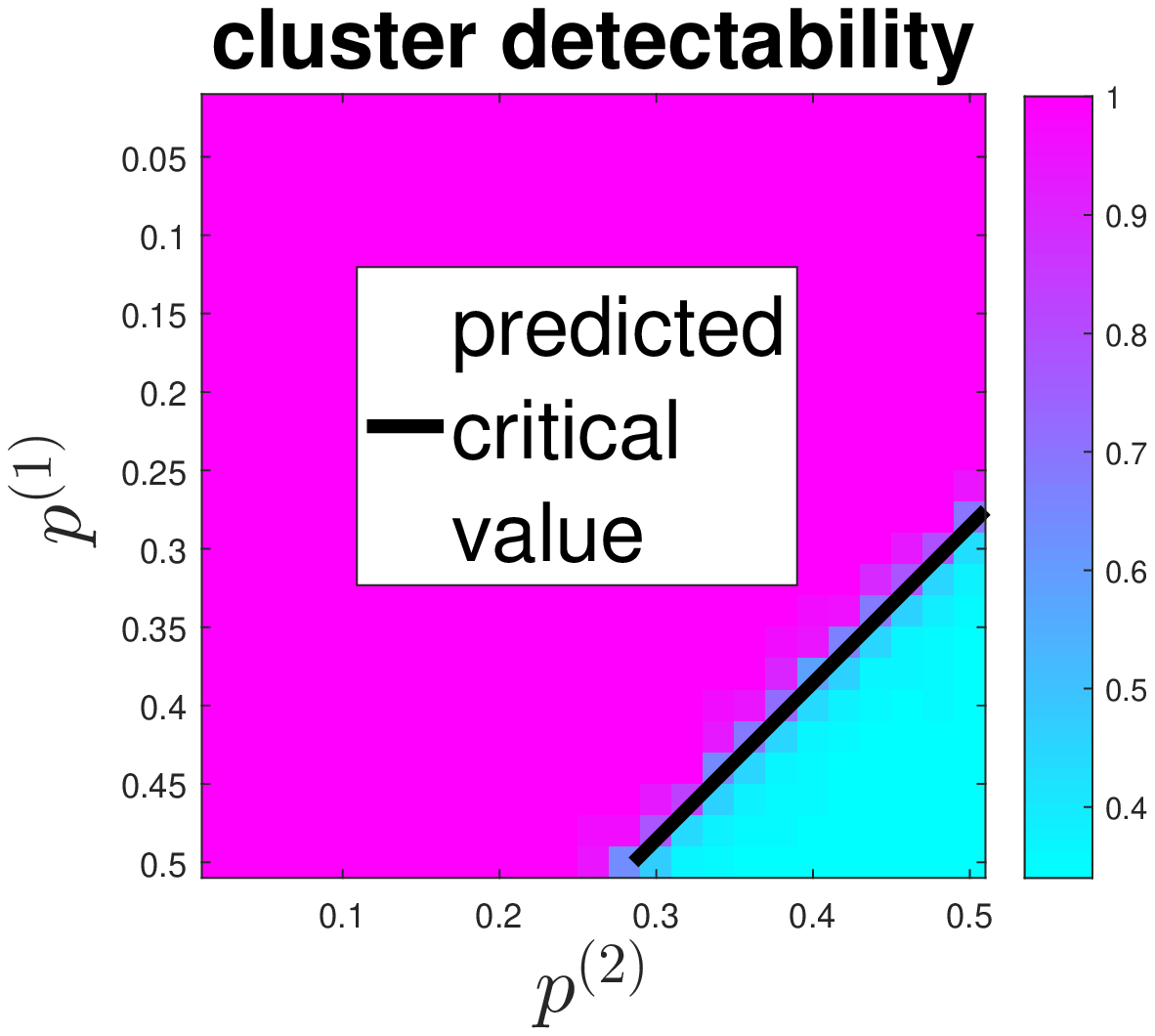}
		\end{subfigure}		
		\centering
		\begin{subfigure}[b]{0.245\linewidth}
			\includegraphics[width=\textwidth]{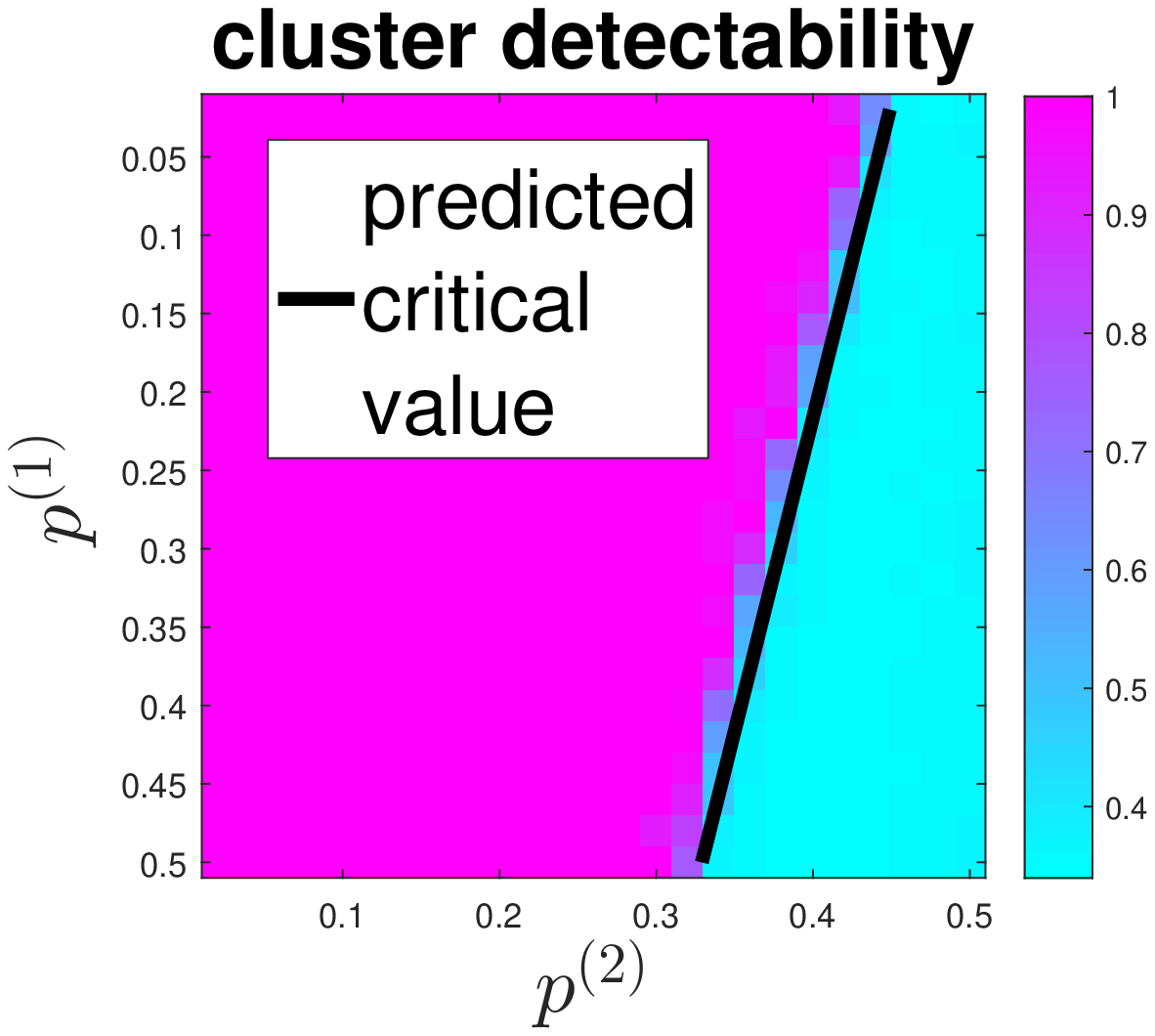}
		\end{subfigure}		
		\begin{subfigure}[b]{0.24\linewidth}
			\includegraphics[width=\textwidth]{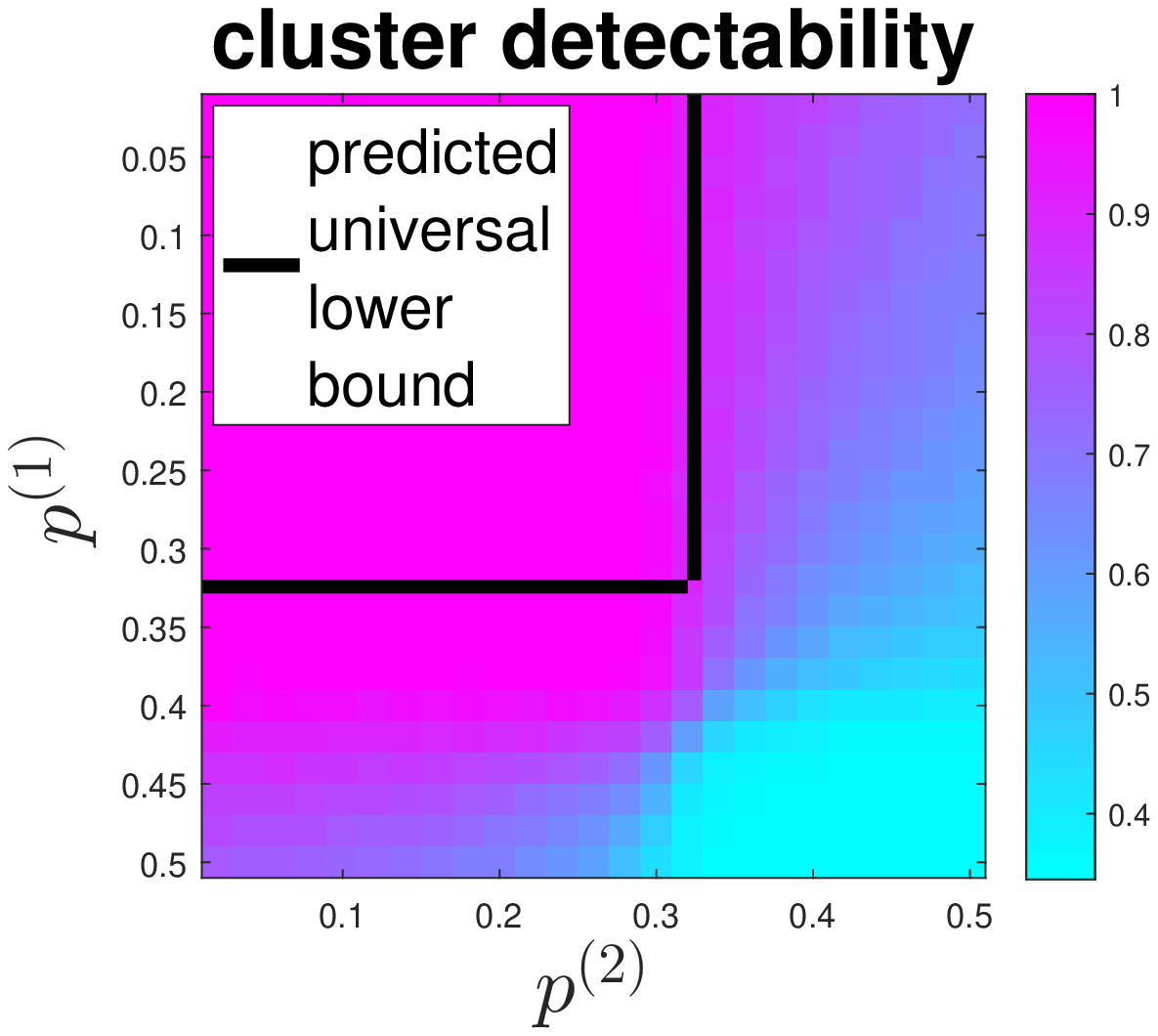}
		\end{subfigure}					
		\vspace{-6mm}
		\caption{Phase transitions in the accuracy of multilayer SGC with respect to different layer weight vector $\bw=[w_1~w_2]^T$ for the two-layer correlated graph model, where		
			 $n_1=n_2=n_3=1000$, $q_{11}=0.3$, $q_{10}=0.2$, $q_{01}=0.1$, and $q_{00}=0.4$.
			 	 From left to right, $(w_1,w_2)=(0.8,0.2),~(0.5,0.5),$ and $(0.2,0.8)$, 	respectively. The last plot is the geometric mean, where $w_1$ is uniformly drawn from $[0,1]$ with unit interval $0.1$.  			 
			  The results are averaged over 10 runs. 
			}
		\label{Fig_two_layer_detectability}
				\vspace{-3mm}
	\end{figure}

By varying the noise level $\{\pl\}_{\ell=1}^2$, Fig. \ref{Fig_two_layer_detectability} shows the accuracy of multilayer SGC with respect to different layer weight vector $\bw=[w_1~w_2]^T$, where the accuracy is evaluated in terms of cluster detectability, i.e., the fraction of correctly identified nodes in the same cluster. Given a fixed $\bw$, as proved in Theorem \ref{thm_spec_ML}, there is indeed a phase transition in cluster detectability that separates the noise level $\{\pl\}_{\ell=1}^2$ into two regimes: a reliable regime where high clustering accuracy is guaranteed, and an unreliable regime where high clustering accuracy is impossible. Furthermore, the critical value of $\{\pl\}_{\ell=1}^2$ that separates these two regimes are successfully predicted by Theorem \ref{thm_spec_ML} (c), which validates the phase transition analysis. The rightmost plot in Fig. \ref{Fig_two_layer_detectability} shows the geometric mean of cluster detectability from different layer weight vectors. There is an universal region of perfect cluster detectability that includes the region specified by the universal phase transition lower bound in (\ref{eqn_LB_tLB}). 

\begin{figure}[t]
	\centering
	\begin{subfigure}[b]{0.245\linewidth}
		\includegraphics[width=\textwidth]{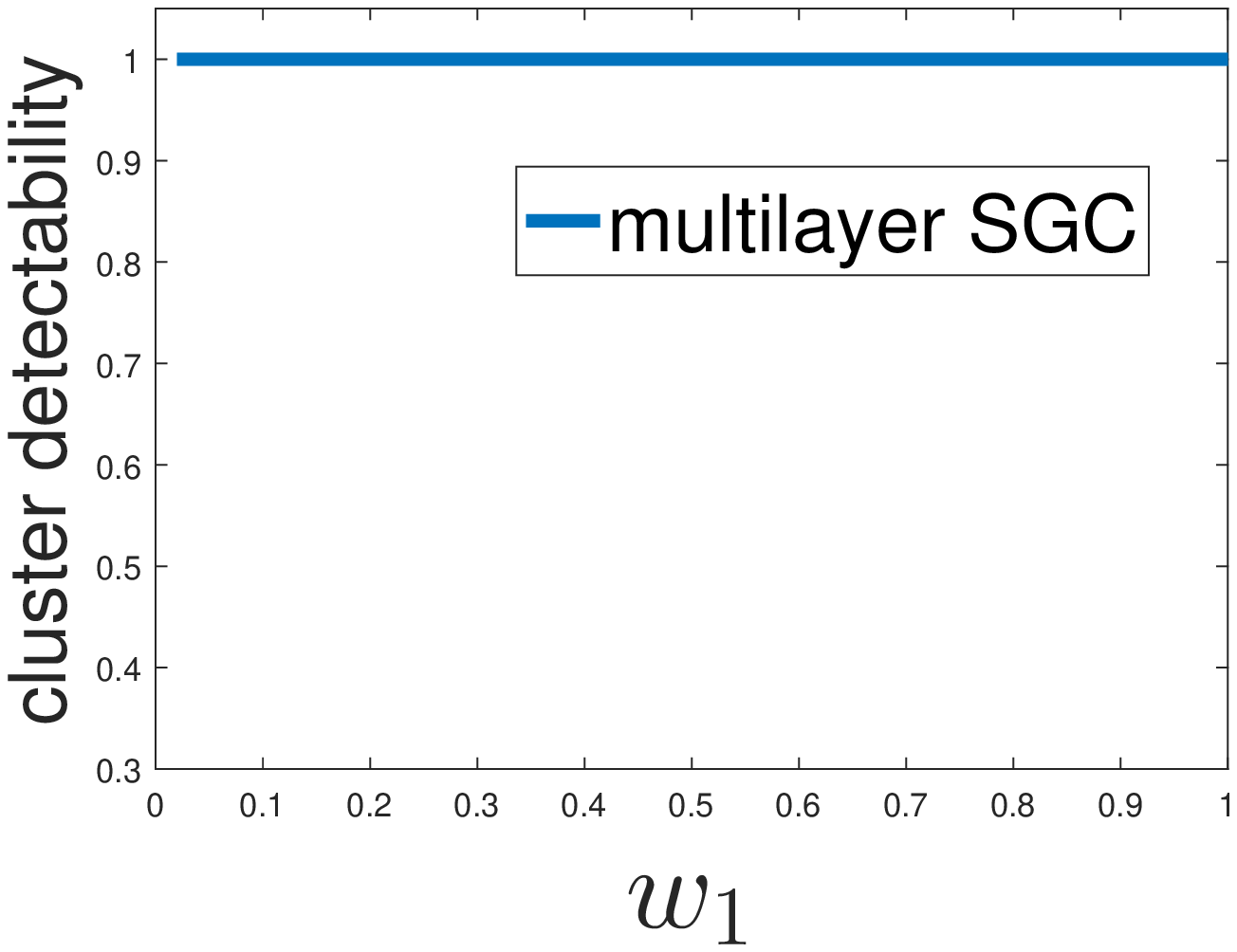}
	\end{subfigure}%
	\centering
	\begin{subfigure}[b]{0.245\linewidth}
		\includegraphics[width=\textwidth]{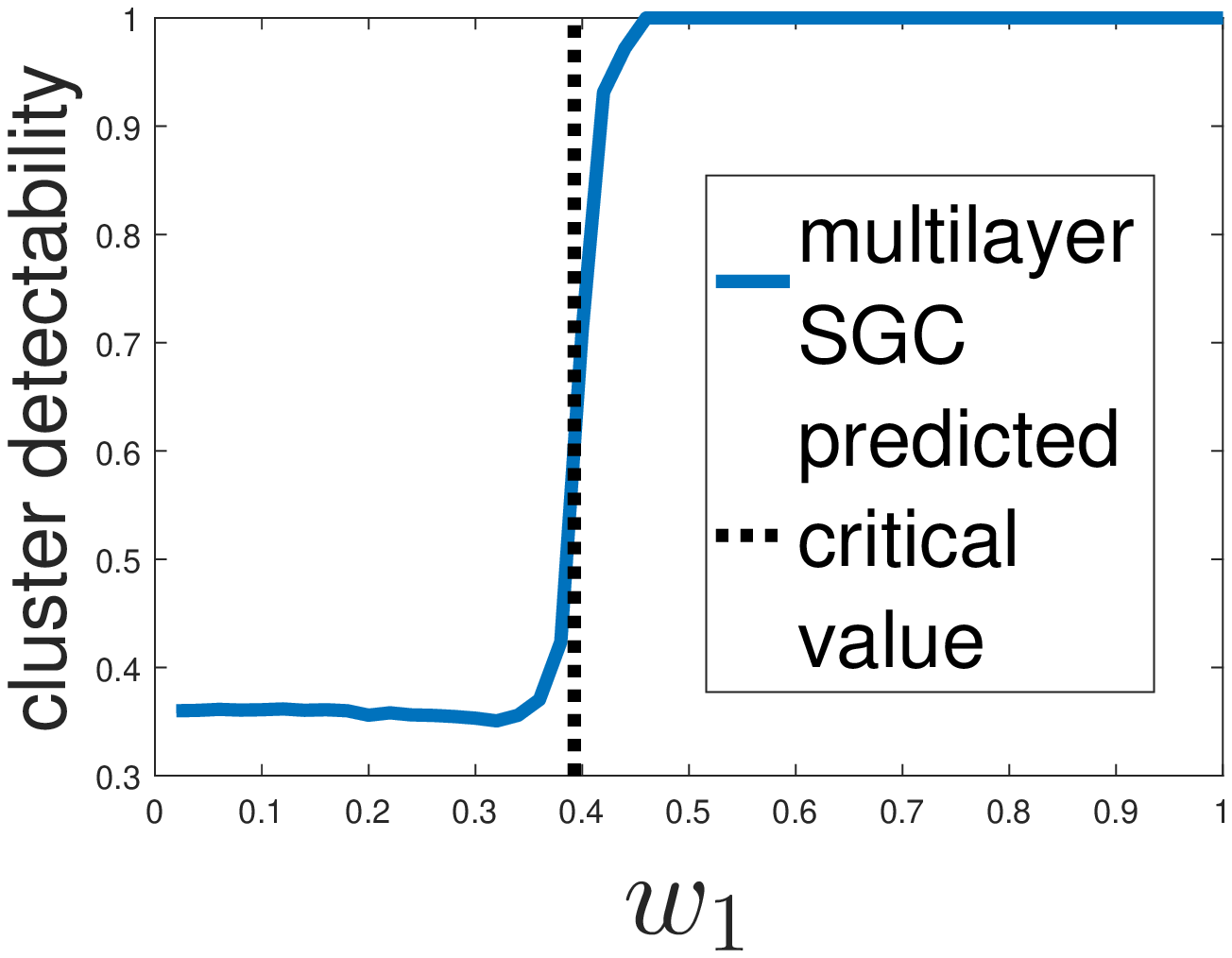}
	\end{subfigure}
	\centering
	\begin{subfigure}[b]{0.245\linewidth}
		\includegraphics[width=\textwidth]{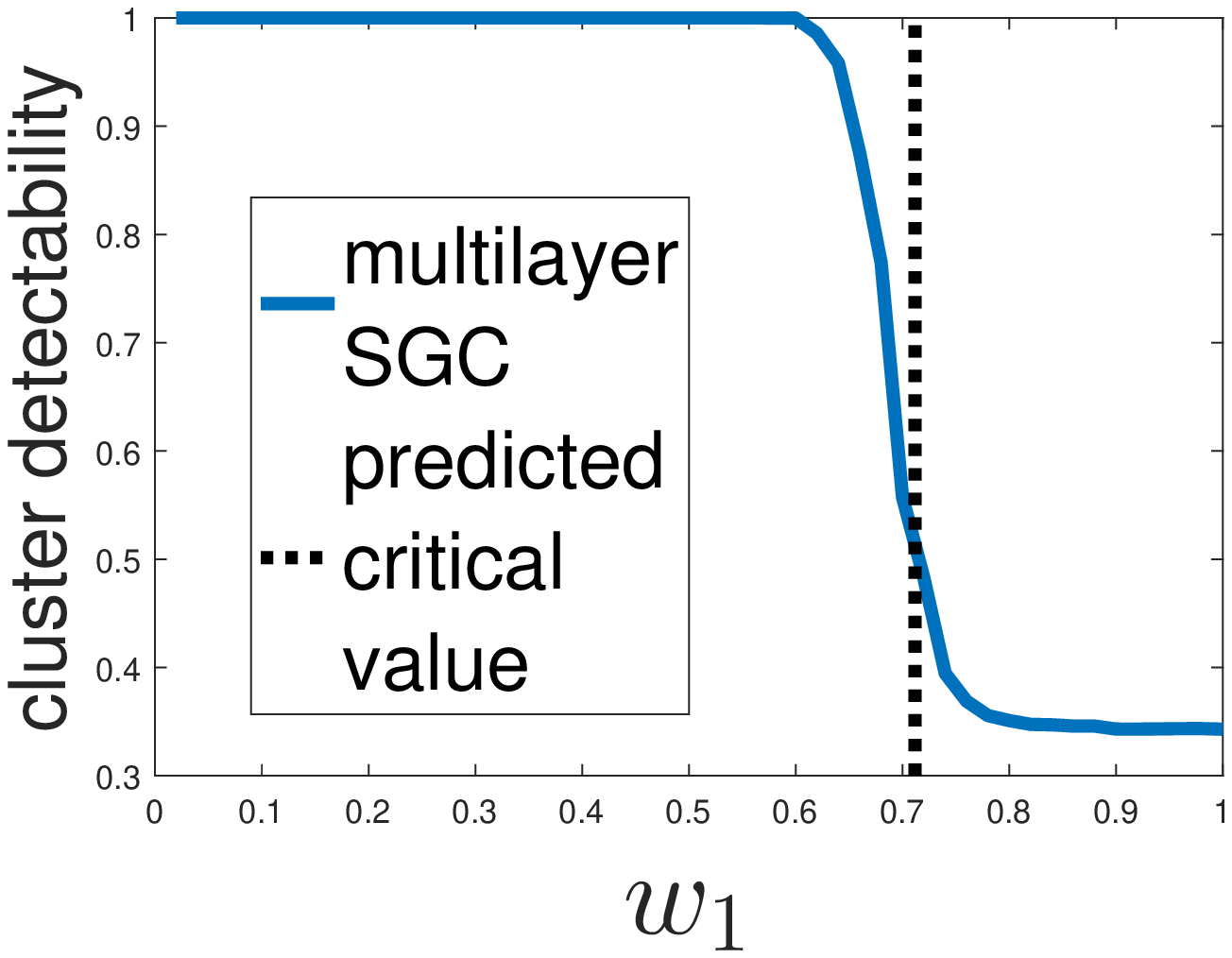}
	\end{subfigure}
	\centering
	\begin{subfigure}[b]{0.245\linewidth}
		\includegraphics[width=\textwidth]{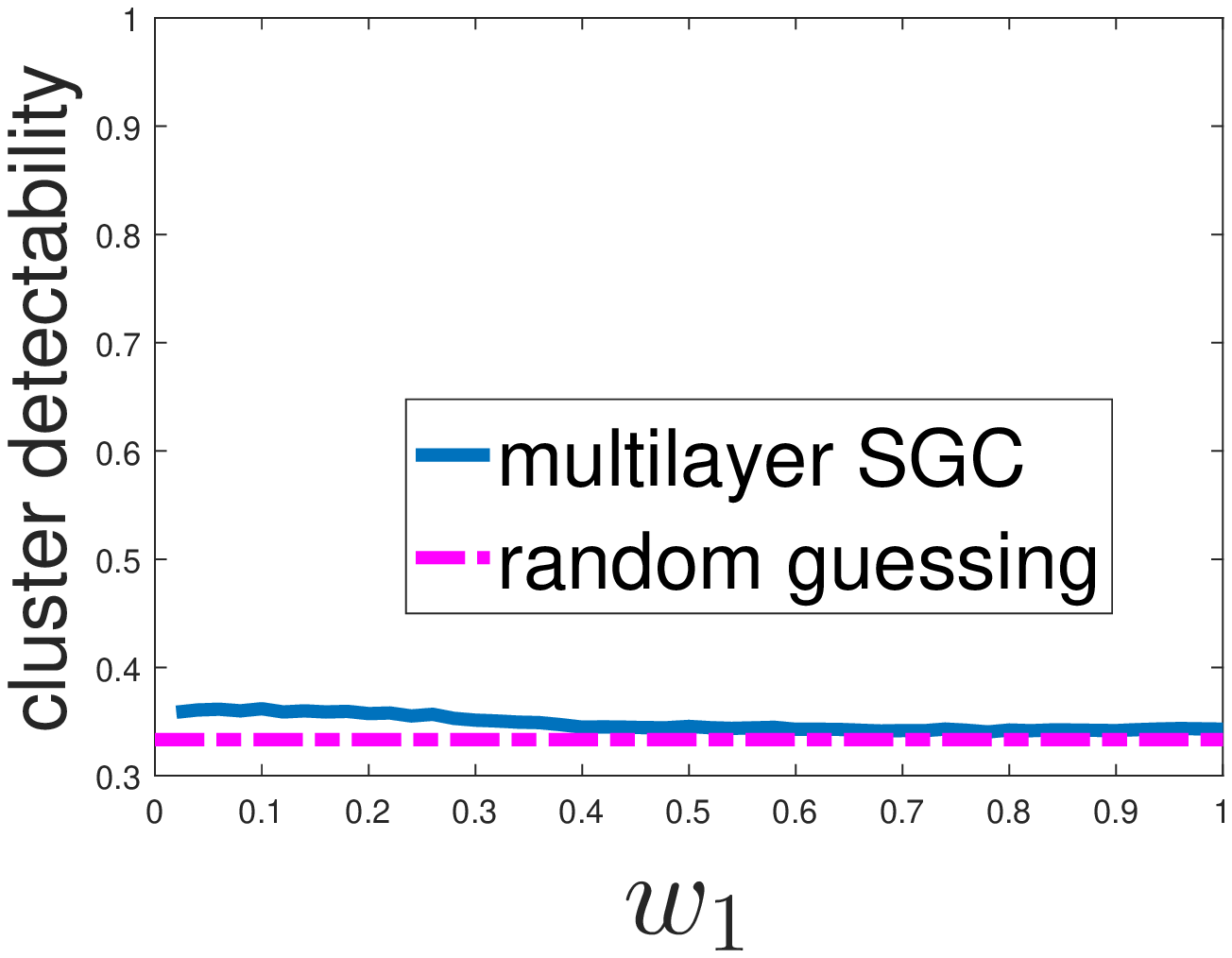}
	\end{subfigure}		
			\vspace{-6mm}				
	\caption{The effect of the layer weight vector $\bw=[w_1~w_2]^T$ on the accuracy of multilayer SGC with respect to difference noise level $\{ \pl\}_{\ell=1}^2$ for the same two-layer correlated graph model as in Fig. \ref{Fig_two_layer_detectability}. 
		From left to right, $(p^{(1)},p^{(2)})=(0.2,0.2),~(0.2,0.5),~(0.5,0.2),$ and $(0.5,0.5)$, respectively.
		 The results are averaged over 50 runs. 	
	} 
	\label{Fig_two_layer_weight} 
	\vspace{-4mm}
\end{figure}

\subsection{Phase transitions incurred by layer weights}
Next we investigate the effect of layer weight vector $\bw$ on multilayer SGC via convex layer aggregation given fixed noise levels.
In the two-layer graph setting, since by definition $w_2=1-w_1$, it suffices to study the effect of $w_1$ on clustering accuracy.
Fig. \ref{Fig_two_layer_weight} shows the clustering accuracy by varying $w_1$ under the two-layer correlated graph model. As shown in Fig. \ref{Fig_two_layer_weight}, if each layer has low noise level (left plot), then any layer weight vector $\bw \in \cW_2$ can lead to correct clustering result. If one layer has high noise level (middle plots), then there exists a critical value $w_1^\star \in [0,1]$ that separates the cluster detectability into a reliable regime and an unreliable regime. In particular, Theorem \ref{thm_spec_ML} implies that the critical value $w_1^\star$, if existed, satisfies the condition $t^{\bw}= t^{\bw^*}$ when $\bw=[w_1^\star,1-w_1^\star]^T=\bw^*$, which is equivalent to
\begin{align}
\label{eqn_weight_equivalent}
&\frac{K-1}{K} \Lb w_1^\star p^{(1)} + (1-w_1^\star) p^{(2)}  \Rb= w_1^\star \cdot \min_{k \in \{1,2,\ldots,K\}} \SK \lb \frac{\bL^{(1)}_k}{n} \rb  \nonumber \\
&~~~+ (1-w_1^\star) \cdot \min_{k \in \{1,2,\ldots,K\}} \SK \lb \frac{\bL^{(2)}_k}{n} \rb. 
\end{align}
It is observed that the empirical critical value $w_1^\star$ matches the predicted value from   (\ref{eqn_weight_equivalent}). Lastly, if each layer has high noise level (right plot), then no layer weight vector can lead to correct clustering result, and the corresponding cluster detectability is similar to random guessing of clustering accuracy $\frac{1}{K} \approx$ 33.33\%.

\section{Conclusion}
\label{sec_conclusion}
This paper establishes a phase transition analysis on multilayer spectral graph clustering (SGC) via convex layer aggregation under a novel multilayer signal plus noise model. By varying the noise level, we specify the critical value that separates the clustering performance of multilayer  (SGC) into a reliable regime and an unreliable regime. Numerical experiments validate the phase transitions incurred by noise levels and layer weights, which are successfully predicted by the developed analytical results.

\clearpage
\bibliographystyle{IEEEtran}
\bibliography{IEEEabrv20160824,CPY_ref_20160824}

\end{document}

%% file: notation_20160812.tex
\newcommand{\ra}{\rightarrow}
\newcommand{\LB}{\left\{}
\newcommand{\RB}{\right\}}
\newcommand{\Lb}{\left[}
\newcommand{\Rb}{\right]}
\newcommand{\lb}{\left(}
\newcommand{\rb}{\right)}
\newcommand{\trace}{\textnormal{trace}}
\newcommand{\diag}{\textnormal{diag}}

\newcommand{\nmin}{{n_{\min}}}
\newcommand{\nmax}{{n_{\max}}}

\newcommand{\tUBw}{t_{\text{UB}}^{\mathbf{w}}}
\newcommand{\tLBw}{t_{\text{LB}}^{\mathbf{w}}}
\newcommand{\tw}{t^{\mathbf{w}}}
\newcommand{\twstar}{{t^{\mathbf{w}}}^*}
\newcommand{\cwstar}{{c^{\mathbf{w}}}^*}

\newcommand{\pl}{p^{(\ell)}}
\newcommand{\tl}{t^{(\ell)}}
\newcommand{\twmax}{t^{\mathbf{w}}_{\max}}

\newcommand{\tlmax}{t^{(\ell)}_{\max}}

\newcommand{\pijl}{p_{ij}^{(\ell)}}
\newcommand{\tijl}{t_{ij}^{(\ell)}}

\newcommand{\bone}{\mathbf{1}}
\newcommand{\bzero}{\mathbf{0}}

\newcommand{\bw}{\mathbf{w}}

\newcommand{\bsw}{\mathbf{s}^\bw}

\newcommand{\bv}{\mathbf{v}}

\newcommand{\Wbarijl}{\overline{W}_{ij}^{(\ell)}}
\newcommand{\Wbarl}{\overline{W}^{(\ell)}}

\newcommand{\SK}{S_{2:K}}
\newcommand{\bX}{\mathbf{X}}
\newcommand{\bY}{\mathbf{Y}}

\newcommand{\bI}{\mathbf{I}}

\newcommand{\btLw}{\widetilde{\mathbf{L}}^\bw}
\newcommand{\btY}{\widetilde{\mathbf{Y}}}

\newcommand{\bAl}{\mathbf{A}^{(\ell)}}
\newcommand{\bAw}{\mathbf{A}^{\bw}}
\newcommand{\bV}{\mathbf{V}}

\newcommand{\bWl}{\mathbf{W}^{(\ell)}}
\newcommand{\bWw}{\mathbf{W}^{\bw}}

\newcommand{\bFl}{\mathbf{F}^{(\ell)}}

\newcommand{\bM}{\mathbf{M}}

\newcommand{\bL}{\mathbf{L}}
\newcommand{\bLw}{\mathbf{L}^\bw}
\newcommand{\bLl}{\mathbf{L}^{(\ell)}}

\newcommand{\bCl}{\mathbf{C}^{(\ell)}}

\newcommand{\bSw}{\mathbf{S}^\bw}
\newcommand{\bSl}{\mathbf{S}^{(\ell)}}

\newcommand{\cW}{\mathcal{W}}

\newcommand{\cV}{\mathcal{V}}
\newcommand{\cE}{\mathcal{E}}

\newcommand{\bbR}{\mathbb{R}}